\documentclass[letterpaper, 10 pt, conference]{ieeeconf}

\IEEEoverridecommandlockouts                              

\usepackage[nolist,nohyperlinks]{acronym}

\usepackage[style=ieee,citestyle=numeric-comp,maxbibnames=99]{biblatex}
\usepackage[urlcolor=cyan,colorlinks=false]{hyperref}
\newcommand\rurl[1]{%
  \href{http://#1}{\nolinkurl{#1}}%
}
\newcommand{\ra}[1]{\renewcommand{\arraystretch}{#1}}
\usepackage{booktabs}

\usepackage[at]{easylist}

\usepackage{xcolor}

\usepackage{amsfonts}
\usepackage{amssymb}
\usepackage{amsmath}

\newcommand{\vect}[1]{\mathbf{#1}}
\newcommand{\mat}[1]{\mathbf{#1}}
\newcommand{\image}[1]{\mathcal{#1}}

\usepackage{graphicx}
\usepackage{subcaption}

\usepackage{xspace}

\newcommand{\reffig}[1]{Fig.~\ref{#1}}

\newcommand{\etal}{\textit{et al.}\xspace}

\begin{document}

\title{\LARGE \bf nvblox: GPU-Accelerated Incremental Signed Distance Field Mapping}


\author{\authorblockN{Alexander Millane\authorrefmark{1}\authorrefmark{2}, Helen Oleynikova\authorrefmark{1}\authorrefmark{3}, Emilie Wirbel\authorrefmark{2}, Remo Steiner\authorrefmark{2},\\ Vikram Ramasamy\authorrefmark{2}, David Tingdahl\authorrefmark{2}, Roland Siegwart\authorrefmark{3}}
\authorblockA{\authorrefmark{1}Equal contribution}
\authorrefmark{2}NVIDIA, Switzerland, \authorrefmark{3}Autonomous Systems Lab, ETH Z{\"u}rich, Switzerland\vspace{-0.1cm}
}

\begin{acronym}
\acro{SDF}{Signed Distance Field}
\acro{TSDF}{Truncated Signed Distance Field}
\acro{ESDF}{Euclidean Signed Distance Field}
\acro{SLAM}{Simultaneous Localization and Mapping}
\acro{FLOPs}{floating point operations per second}
\acro{NERF}{neural radiance field}
\acro{FoV}{Field of View}
\acro{PBA}{Parallel Banding Algorithm}
\end{acronym}

\urlstyle{tt}

\maketitle

\begin{abstract}
Dense, volumetric maps are essential to enable robot navigation and interaction with the environment. 
To achieve low latency, dense maps are typically computed on-board the robot, often on computationally constrained hardware. 
Previous works leave a gap between CPU-based systems for robotic mapping which, due to computation constraints, limit map resolution or scale, and GPU-based reconstruction systems which omit features that are critical to robotic path planning, such as computation of the \ac{ESDF}. 
We introduce a library, \textit{nvblox}, that aims to fill this gap, by GPU-accelerating robotic volumetric mapping. 
\textit{Nvblox} delivers a significant performance improvement over the state of the art, achieving up to a $177\times$ speed-up in surface reconstruction, and up to a $31\times$ improvement in distance field computation, and is available open-source\footnotemark[1].
\end{abstract}

\IEEEpeerreviewmaketitle


\section{Introduction}

To navigate and interact with their environment, robots typically build an internal representation of the world. 
Significant research in the past decades~\cite{cadena2016past} has focused on building maps that are both useful for robotic path-planning, and efficient to construct. However, fulfilling these two requirements simultaneously remains challenging.

Various successful systems have emerged for solving the \ac{SLAM} problem efficiently~\cite{mur2015orb, schneider2018maplab}. 
Typically these systems build sparse representations of the environment in order to reach real-time rates. 
Sparse maps have proven effective for localization, however, navigation also requires dense obstacle information. 

Several systems have emerged for building denser representations of the environment on the CPU~\cite{oleynikova2017voxblox, hornung2013octomap, reijgwart2019voxgraph} that are suitable for robotic path-planning. 
However, the frequency, resolution, and scale at which these systems can operate is limited by the computational burden of 3D reconstruction on a CPU.
To address these limitations, systems utilizing GPU programming have emerged~\cite{izadi2011kinectfusion, whelan2015elasticfusion}.
These systems, however, have typically focused on reconstruction alone and have omitted features needed in a robotic path-planning context, such as incremental computation of the signed distance field and its gradients, as well as an explicit representation of free space. 
We aim to address this gap.

\begin{figure}
    \centering
    \includegraphics[width=1.0\columnwidth]{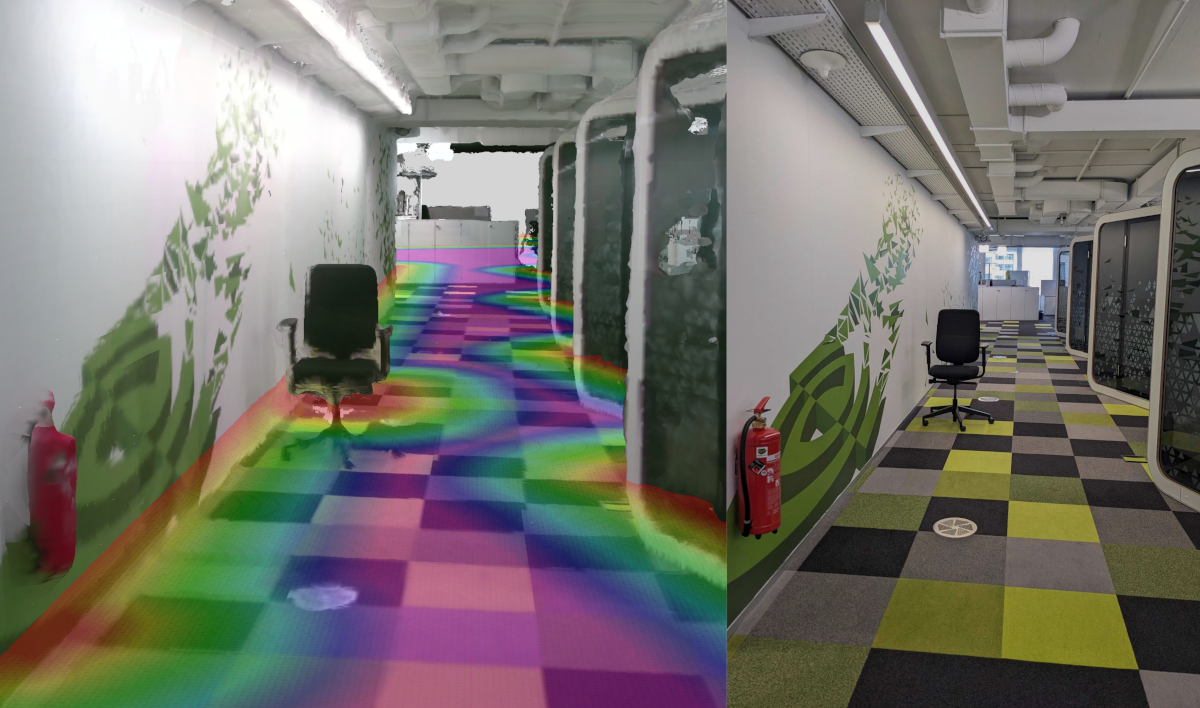}
    \caption{An example of an \textit{nvblox} reconstruction in an office environment (left) compared to a photo of the mapped scene (right). The reconstruction is built in real-time from a handheld Intel RealSense Depth Camera D455. It shows the mesh and a slice through the distance field. All operations to build this reconstruction are performed in real-time on an embedded GPU.}
    \label{fig:teaser}
    \vspace{-5mm}
\end{figure}

This paper introduces \textit{nvblox}, a library for volumetric mapping on the GPU, specifically targeted at robotic path-planning.
\textit{Nvblox} produces high-resolution surface reconstructions at real-time rates, even on embedded GPUs. 
In addition \textit{nvblox} also produces distance fields, which are a key output for planning collision-free paths.
Central to our approach is the use of parallel computation on the GPU for all aspects of the pipeline, including queries.
We show the efficacy and efficiency of \textit{nvblox} on several public datasets, and applied to several robotic use-cases, such as path planning for robot arms, flying robots, ground robots, and for mapping of dynamic obstacles such as people.

In summary, the contribution of this paper is a GPU-accelerated \ac{SDF} library with a convenient and flexible interface. \textit{Nvblox} fuses in data from RGB-D sensors and/or LiDAR, achieving up to $177\times$ faster surface reconstruction (\ac{TSDF}) and $31\times$ faster distance field (\ac{ESDF}) computation than state-of-the-art CPU-based methods~\cite{han2019fiesta, oleynikova2017voxblox}.
The library is made available open-source\footnote{\rurl{github.com/nvidia-isaac/nvblox}} in both ROS1\footnote{\rurl{github.com/ethz-asl/nvblox_ros1}} and ROS2\footnote{\rurl{github.com/NVIDIA-ISAAC-ROS/isaac_ros_nvblox}}.

\section{Related Work}
Mapping is a well-explored problem in robotics~\cite{cadena2016past}.
We can categorize robotic mapping approaches into two broad categories: sparse and dense. 
Sparse methods focus on creating a map representation for pose estimation and localization while dense methods focus on reconstructing the geometry of the environment.

Systems for dense mapping can be organized by the underlying representation of the environment.
LSD-SLAM~\cite{engel2014lsd} and DVO-SLAM~\cite{kerl2013dense} build a map consisting of keyframe depth-maps. 
Kintinuous~\cite{whelan2012kintinuous} and ElasticFusion~\cite{whelan2015elasticfusion} build reconstructions in the form of a deformable mesh and collection of surfels, respectively. 
More recently, Kimera~\cite{rosinol2020kimera} builds a semantic mesh of the environment. 
These approaches build visually compelling reconstructions, but are difficult to use for robotic path-planning, as they lack information about observed free space and focus only on surfaces.

Recently, reconstruction systems based on \acp{NERF} have gained significant attention~\cite{mildenhall2021nerf}.
The original offline approach has seen dramatic speedups~\cite{muller2022instant}, leading to implementations that generate reconstructions in real-time~\cite{sucar2021imap, zhu2022nice}.
However, these approaches heavily keyframe the input image data and are therefore unlikely to be reactive enough for robot path-planning in the control loop.


Voxel-based methods build reconstructions that are well-suited to robot path-planning tasks. 
Voxels capture the reconstructed quantity (for example occupancy probability) over the \textit{volume} of 3D space, and can therefore represent free-space, not only surfaces.
The most common approach to volumetric reconstruction is occupancy grid mapping~\cite{thrun2002probabilistic} and its efficient implementation in 3D, Octomap~\cite{hornung2013octomap}.
These approaches are widely used in robotic mapping and are the default in common robotics toolkits~\cite{macenski2020marathon, macenski2020spatio}.

Another popular approach to volumetric mapping utilizes a voxelized \ac{TSDF}.
This approach was popularized by KinectFusion~\cite{izadi2011kinectfusion} which generates surface reconstructions using a consumer-grade depth camera.
The original, fixed-grid-based approach was extended to use spatially hashed voxels by Niessner et. al.~\cite{niessner2013real}.
Voxblox~\cite{oleynikova2017voxblox} follows the approach of voxel-hashing but adds \ac{ESDF} computation, a feature of particular importance for robotic path-planning. 
Voxblox has been used in many follow-up works which have used it for planning~\cite{tranzatto2022cerberus, dang2020graph}, as well as extended its mapping capabilities, for example for global mapping~\cite{reijgwart2019voxgraph} and semantic mapping~\cite{grinvald2019volumetric}.
Despite its success, voxblox is limited in the resolution of maps it builds due to the computational complexity of updating a high-resolution voxel grid.

Several works have focused on decreasing voxblox's ESDF generation error and runtime.
Voxfield~\cite{pan2022voxfield}, for example, removes the inaccuracies caused by voxblox's quasi-Euclidean distance estimation and improves the ESDF runtime by up to 42\%.
Similarly, FIESTA speeds up voxblox's ESDF computation by $4\times$ while also computing full Euclidean distance.
\textit{Nvblox} also uses full Euclidean distance, therefore reducing the ESDF error, but is on average $31\times$ faster than voxblox.

By improving existing methods through GPU acceleration, we create a library that provides a suitable representation for a large body of path planners and other downstream applications, while reducing the runtime and allowing the creation of higher-resolution maps on the robot.


\section{Problem Statement}

Given a sequence of measurements from an RGB-D camera and/or a LiDAR, we aim to build a volumetric reconstruction of the scene.
In particular, we compute the surface reconstruction (expressed as either occupancy or the \ac{TSDF}) and the distance field (expressed as the \ac{ESDF}). 
Our reconstructions are functions $\Phi: \mathbb{R}^3 \to \phi$, which maps a point in 3D space to some mapped quantity $\phi$, for example, distance, occupancy, or color.
These functions are voxelized, i.e. represented as a sparse set of samples on a regular 3D grid, where samples are allocated in regions of space that have been observed by the sensor. 
We assume that at time step $i$ the sensor frame $C^i$ is localized in a frame $L$ such that we have access to the sensor pose $\mat{T}_{LC^i} \in \text{SE}(3)$. 
Observations take the form of depth maps and color images.
A depth map is $\mathcal{D}: \Omega \to d$, where $d \in \mathbb{R}$ is a depth value in meters. 
The domain $\Omega$ is the image plane in the case of depth cameras, and the beam angles in the case of rotating LiDARs.
Similarly, color images are $\mathcal{C}: \Omega \to c$ where $c \in \mathbb{R}^3$.
In the remainder of this paper, we will refer to observations from both camera and LiDAR as images and treat the two equally.

\section{System architecture description}
\label{sec:system_architecture}

\begin{figure}
    \centering
    \includegraphics[width=0.95\columnwidth]{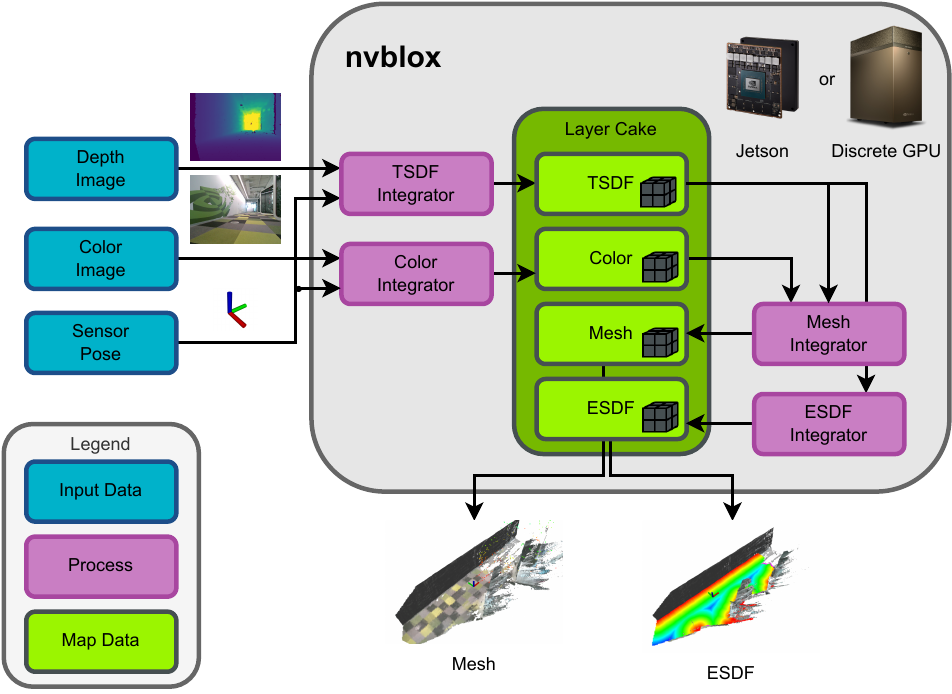}
    \caption{The system architecture of \textit{nvblox}, configured for \ac{TSDF} mapping from an RGB-D camera sensor. The reconstructed map (called a \textit{LayerCake}) is composed of co-located and aligned 3D voxel grids. Input depth-maps and color images are integrated into the \ac{TSDF} and Color voxel layers, from which voxel grids containing the \ac{ESDF} and a mesh reconstruction are derived. See Sec.~\ref{sec:system_architecture} for details.}
    \label{fig:system_architecture}
    \vspace{-5mm}
\end{figure}

The architecture of \textit{nvblox} is shown in Fig.~\ref{fig:system_architecture}. 
The system consists of multiple components: the reconstructed map, which contains several \textit{Layers}, \textit{Integrators} that add sensor data to the map and components that transform one layer type into another, such as mesh and \ac{ESDF} integrators.
We discuss these components in more detail below.

\subsection{The Map}
\label{sec:map}

Our reconstruction is represented as several overlapping 3D voxel grids, called \textit{Layers}, following \cite{oleynikova2017voxblox}. 
Each \textit{Layer} of the map stores a different type of (user-defined) data for overlapping aligned volumes of 3D space. 
The map is sparse, such that voxels are only allocated in regions of 3D space that are observed during mapping.  
This sparsity is achieved using a two-level hierarchy, following~\cite{niessner2013real}. 
The first level is a hash table that maps 3D grid indices to \textit{VoxelBlocks}. 
In \textit{nvblox} this hash table can be queried in GPU kernels using an interface based on stdgpu~\cite{stotko2019stdgpu}. 
In the second level, each \textit{VoxelBlock} contains a $8\! \times \! 8 \! \times \! 8$ group of densely allocated voxels which are stored contiguously in GPU memory, leading to coalesced loads in GPU kernels. 

The map is designed to be extended with new layers. 
To create a new \textit{Layer}, a user needs to specify the contents of a single voxel. 
The library generates the definitions for the corresponding block-hashed voxel grid at compile time, as well as the CPU and GPU interfaces. 
The \textit{nvblox} library includes commonly used \textit{Layers}: \ac{TSDF}, \ac{ESDF}, occupancy, color, and meshes. 

\subsection{Frame Integration}
\label{sec:frame_integration}

Incoming sensor data is added to the reconstruction stored in one of the map layers. 
This occurs in several steps.
We first ray trace through the \textit{VoxelBlock}-grid on the GPU to determine which \textit{VoxelBlocks} are in view using~\cite{amanatides1987fast} and allocate those not yet in the map.
We then project each voxel in view into the depth image:
\begin{equation}
  d = \image{D}[\pi_{\text{sensor}}(\mat{T}_{CL}  \vect{p}_L)]
\end{equation}
where $d \in \mathbb{R}$ is the sampled depth value, $\mat{T}_{CL}$ is the camera pose with respect to the \textit{Layer}, and $\vect{p}_L$ is the voxel center position in the \textit{Layer} coordinate frame.
The sensor projection function for a camera $\pi_{\text{sensor}}$ is defined as
\begin{equation}
    \pi_{\text{camera}}(\vect{p}_C) = 
    \frac{1}{p_{C,z}}
    \left[\begin{array}{ccc}
        f_u & 0 & c_u \\
        0 & f_v & c_v
    \end{array}\right]
    \left[\begin{array}{c}
        p_{C,x} \\
        p_{C,y} \\
        1
    \end{array}\right]
\end{equation}
where $f_u$, $f_v$, $c_u$, $c_v$ are calibrated pinhole camera intrinsics, and for LiDAR
\begin{equation}
    \pi_{\text{lidar}}(\vect{p}_C) = 
    \left[\begin{array}{c}
        (\tan^{-1}(p_{C,y} / p_{C,x}) - \theta_{\text{start}}) * \alpha_{\theta} \\
        (\cos^{-1}(p_{C,z} / r) - \phi_{\text{start}}) * \beta_{\phi}
    \end{array}\right],
\end{equation}
where $\theta_{\text{start}}$, $\phi_{\text{start}}$ are the minimum polar and azimuth angles, and $\alpha_{\theta}$ and $\beta_{\phi}$ are measured in pixels-per-radian and are calculated by dividing the \ac{FoV} by the number of steps/beams in the relevant dimension. The function $\image{D}[\vect{u}]$ indicates sampling the depth image $\image{D}$ at image-coordinates $\vect{u}$. For a camera, we sample using nearest-neighbor, and for LiDAR-based depth images, which can be very sparse, we use linear interpolation with modifications to avoid interpolating over foreground-background boundaries. 

To update voxels, we call a per-voxel update functor on the GPU in parallel over all voxels in view. We will briefly cover two main update functors: one for TSDF maps and one for occupancy maps.
TSDF maps store both a truncated, projective distance ($d_{tsdf}$) and a weight ($w$) per voxel, while occupancy maps store a single log-odds occupancy value $l_{o}$.

The TSDF update functor uses the voxel depth $d_v$, which is the depth from the voxel center to the sensor, to calculate $d_p = d - d_{v}$, the projective distance at that voxel, truncates it to within the positive and negative truncation distance $\epsilon$, 
and computes a weight $w  = f_w(d_p)$,
where $f_w(x)$ is the weighting function, either a constant or based on the sensor's error model (nvblox offers several models). TSDF distance values are combined using a weighted average, and weights are simply added, up to a maximum.

The occupancy update functor, updates a voxel's occupancy probability in log-odds space. A voxel is updated with a constant negative value if $d_p > 0$ (indicating a lower probability of occupancy) or a constant positive value if $d_p < 0$. 
Bayesian fusion in log-odds space is implemented as running addition of the update values (see \cite{thrun2002probabilistic} for details).





For TSDF maps, updated blocks are periodically meshed using a parallelized Marching Cubes~\cite{lorensen1998marching} algorithm.




\subsection{ESDF Computation}
\label{sec:esdf_computation}
The \acf{TSDF} contains projective distances up to a small truncation distance. 
For path planning applications we require \textit{Euclidean} distances 
and for greater distances than the truncation band. 
For a discussion on why a \ac{TSDF} is insufficient for this, we refer the reader to~\cite{oleynikova2016signed}.

Our \acf{ESDF} computation approach has several requirements. It must be both parallelizable on the GPU and incremental, allowing us to update only parts of the map that have changed to reduce computation time. Finally, we avoid simplifying assumptions (like quasi-Euclidean distance used in voxblox~\cite{oleynikova2017voxblox}) to maintain high accuracy of the resulting distance field.


\begin{figure}[t]
    \centering
    \includegraphics[width=1.0\columnwidth]{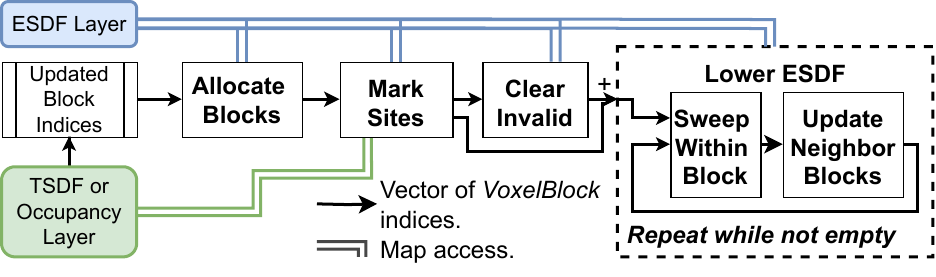}
    \caption{Overall flow of the incremental ESDF computation on the GPU. The algorithm does as much of the update in parallel as possible, iteratively identifying blocks that need to be updated, and then running kernels on all blocks in parallel.} 
    \label{fig:esdf_flow}
    \vspace{-5mm}
\end{figure}

\begin{figure*}[th]
  \centering
\begin{subfigure}[b]{0.35\columnwidth}
    \includegraphics[width=1.0\columnwidth,trim=0 0 0 0 mm, clip=true]{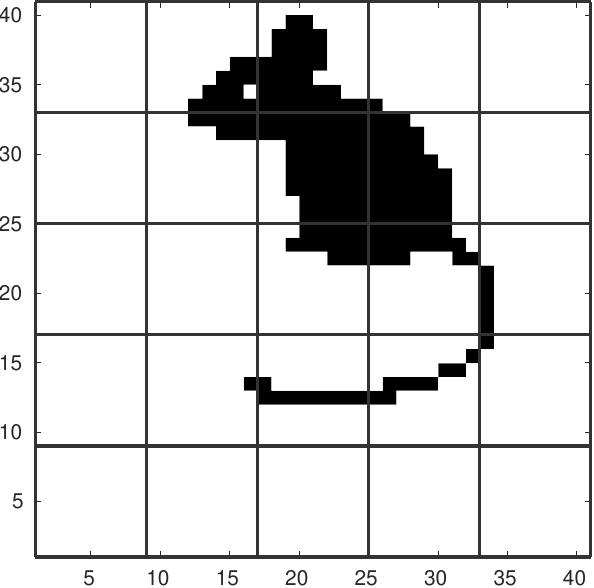}
    \caption{Previous Map}
    \label{fig:cats_previous}
  \end{subfigure}
  \begin{subfigure}[b]{0.35\columnwidth}
    \includegraphics[width=1.0\columnwidth,trim=0 0 0 0 mm, clip=true]{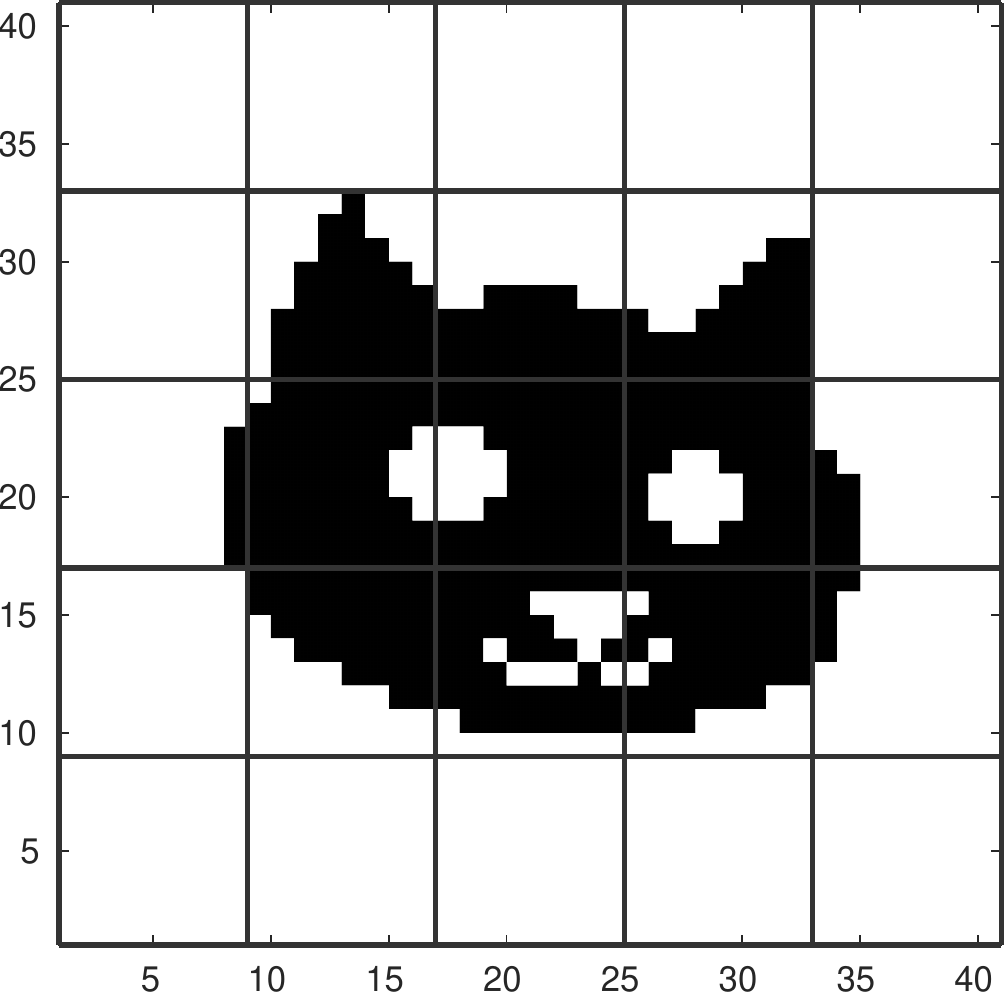}
    \caption{(New) Occupancy}
    \label{fig:cats_occupancy}
  \end{subfigure}
  \begin{subfigure}[b]{0.35\columnwidth}
    \includegraphics[width=1.0\columnwidth,trim=0 0 0 0, clip=true]{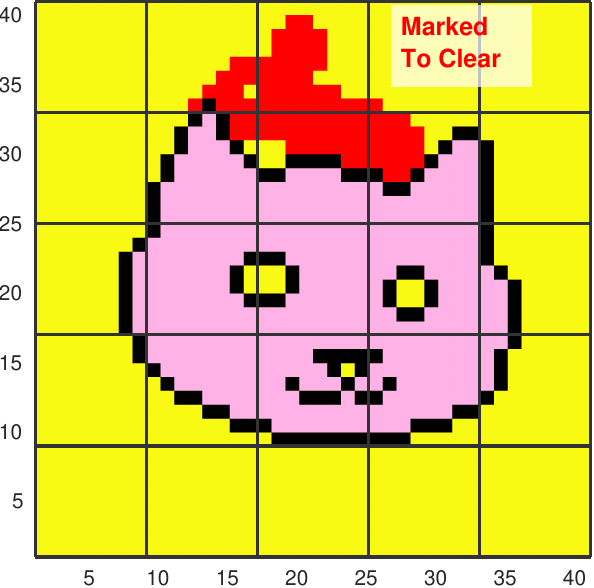}
    \caption{Marking}
    \label{fig:cats_marking}
  \end{subfigure}
\begin{subfigure}[b]{0.35\columnwidth}
    \includegraphics[width=1.0\columnwidth,trim=0 0 0 0, clip=true]{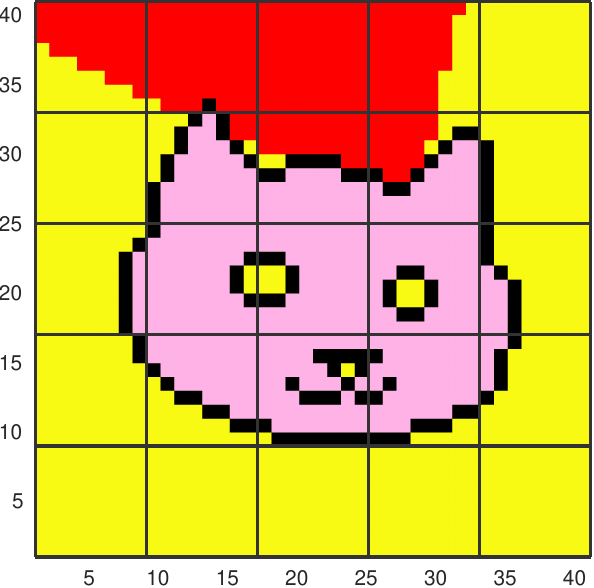}
    \caption{Clearing}
    \label{fig:cats_clearing}
  \end{subfigure}
  \begin{subfigure}[b]{0.35\columnwidth}
    \includegraphics[width=1.0\columnwidth,trim=0 0 0 0, clip=true]{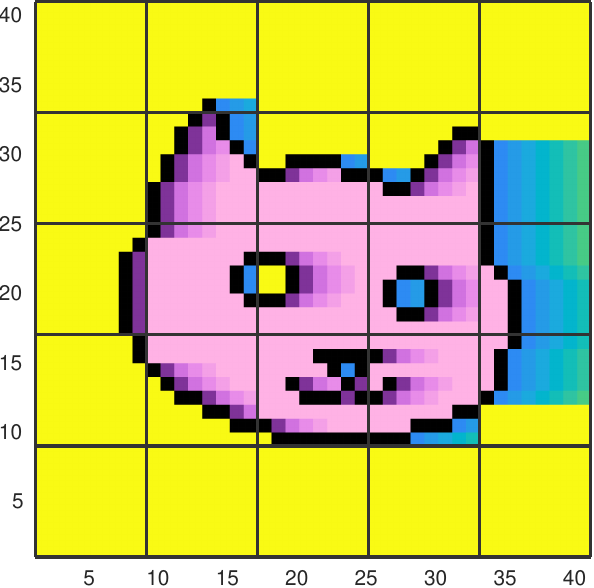}
    \caption{$X+$ Sweep}
    \label{fig:cats_xplus}
  \end{subfigure}
  \vspace{3mm}
    \begin{subfigure}[t][-20pt][b]{0.09\columnwidth}
    \includegraphics[width=1.0\columnwidth,trim=0 0 0 0, clip=true]{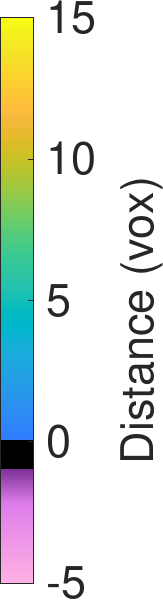}
  \end{subfigure}
  \begin{subfigure}[b]{0.35\columnwidth}
    \includegraphics[width=1.0\columnwidth,trim=0 0 0 0, clip=true]{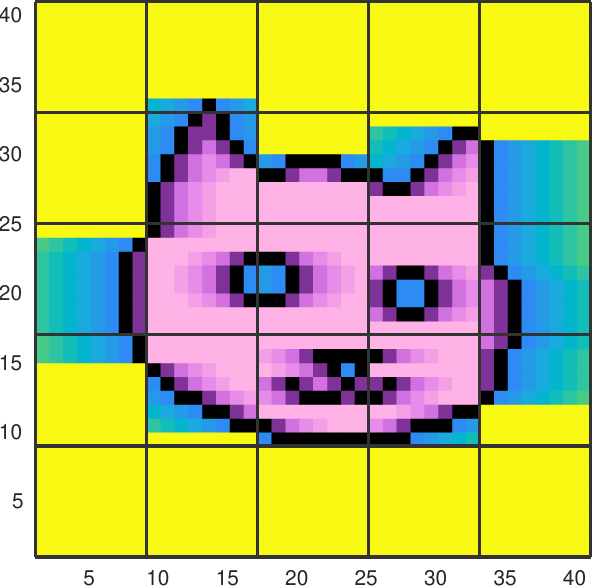}
    \caption{$X-$ Sweep}
    \label{fig:cats_xminus}
  \end{subfigure}
  \begin{subfigure}[b]{0.35\columnwidth}
    \includegraphics[width=1.0\columnwidth,trim=0 0 0 0, clip=true]{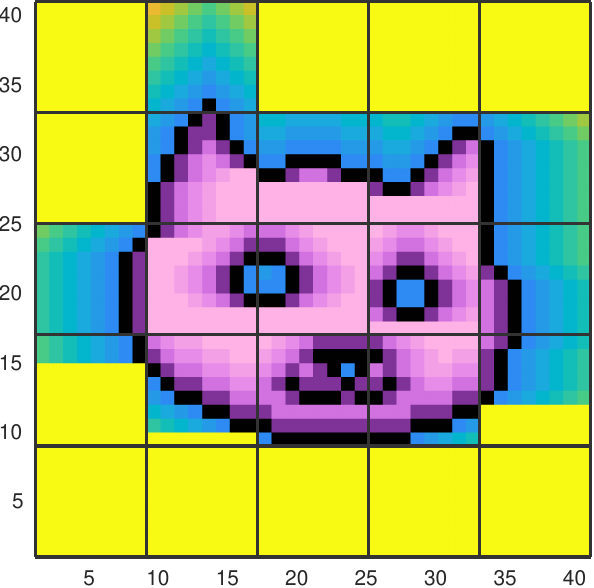}
    \caption{$Y+$ Sweep}
    \label{fig:cats_yplus}
  \end{subfigure}
  \begin{subfigure}[b]{0.35\columnwidth}
    \includegraphics[width=1.0\columnwidth,trim=0 0 0 0, clip=true]{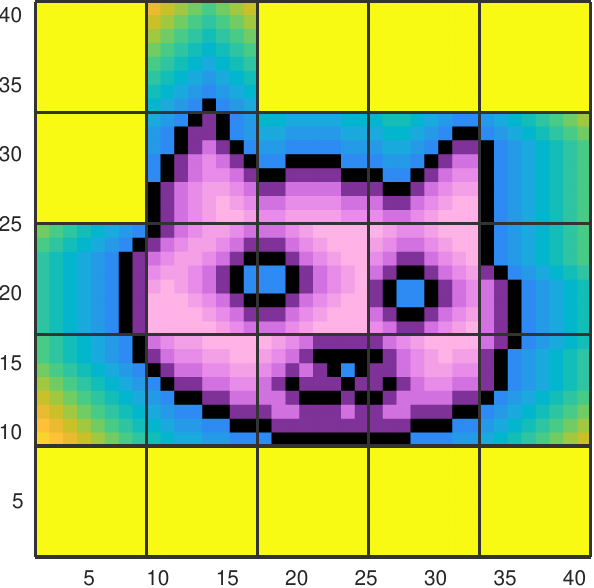}
    \caption{$Y-$ Sweep}
    \label{fig:cats_yminus}
  \end{subfigure}
  \begin{subfigure}[b]{0.35\columnwidth}
    \includegraphics[width=1.0\columnwidth,trim=0 0 0 0, clip=true]{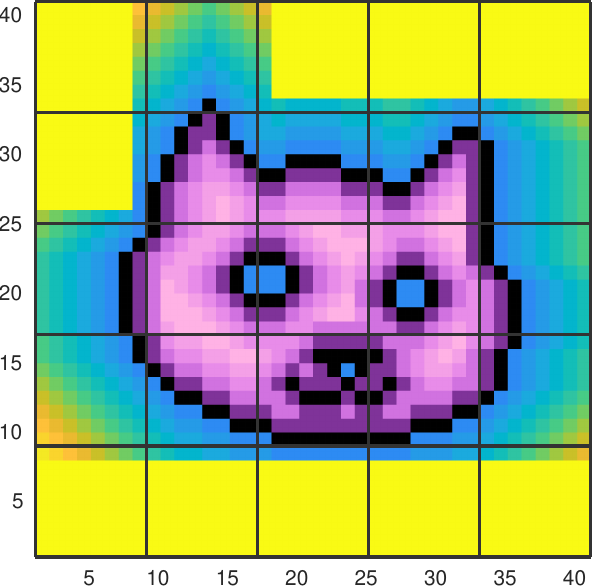}
    \caption{Neighbor Borders}
    \label{fig:cats_border}
  \end{subfigure}
  \begin{subfigure}[b]{0.35\columnwidth}
    \includegraphics[width=1.0\columnwidth,trim=0 0 0 0, clip=true]{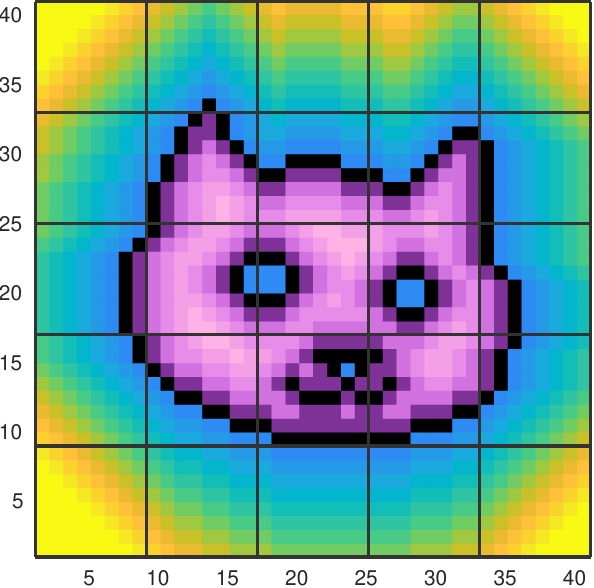}
    \caption{Final Sweeps}
    \label{fig:cats_sweeps}
  \end{subfigure}
  \caption{Marking and lowering, shown step-by-step on an image, for simplicity on 2D occupancy (rather than 3D TSDF). Here we start with a previous map shown in \subref{fig:cats_previous} to demonstrate clearing of previously-occupied space. The bold lines represent \textit{VoxelBlock} boundaries. The general idea of the algorithm is to iteratively compute correct distance values \textit{within} a \textit{VoxelBlock}, and then communicate distance information across \textit{VoxelBlock} boundaries to neighbors.}
  \label{fig:cats}
\end{figure*}

In order to create a highly parallelizeable ESDF computation algorithm, we base our work in spirit on the \ac{PBA} proposed by Cao \etal~\cite{cao2010parallel}. 
The overall flow of the algorithm is shown in \reffig{fig:esdf_flow} and visualized step-by-step in \reffig{fig:cats}.
The general intuition is to keep a list of blocks that need updating, and update all voxels in all affected blocks in parallel, transmit information between blocks, and repeat until convergence. 

\begin{figure}[t]
    \centering
    \includegraphics[width=0.6\columnwidth]{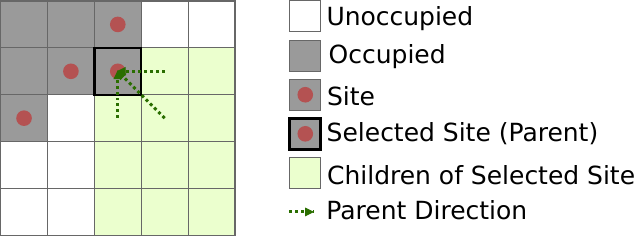}
    \caption{``Sites'' are voxels on the surface boundary, and its children are any voxels which are closer to that voxel than any other site. Each child stores its parent direction - the direction that points to the parent site location (only shown for the closest 3 voxels here for clarity).} 
    \label{fig:parent_child}
    \vspace{-5mm}
\end{figure}

ESDF voxels come in two categories: \textit{sites} and regular voxels. Sites are surface boundary voxels, as shown in \reffig{fig:parent_child}. Site voxels can be \textit{parents} to \textit{child} voxels, and each \textit{child} stores the location its \textit{parent}. Regular voxels can have three states: unknown, free (positive distance), and occupied (negative distance). Each observed voxel also stores its distance to the nearest site.

First, in \textit{Allocate Blocks}, a TSDF layer and a list of updated \textit{VoxelBlocks} are taken as input. Generally, TSDF blocks that were changed since the last iteration of ESDF integration are considered updated (this allows us to run the ESDF update slower than sensor rate). Any new \textit{VoxelBlocks} are allocated in the ESDF as needed. 

The next step is \textit{Mark Sites}. 
We consider voxels to be sites if their TSDF distance is within a threshold $\epsilon$ of the zero crossing (see \reffig{fig:cats_marking}); otherwise, an occupancy voxel is a site if it's adjacent to free voxels.

In addition to marking sites, this function ensures consistency between the ESDF and TSDF or occupancy maps. There are two general cases we need to handle: \textit{(1)} newly occupied (free $\rightarrow$ occupied) or newly observed (unknown $\rightarrow$ either) voxels, and \textit{(2)} newly freed voxels (occupied $\rightarrow$ free).
Newly occupied/observed voxels simply take on their TSDF distance values and are added to \textit{Indices to Update}. Newly free voxels have their distances set invalid. 
To complicate matters, if the cleared voxel was a site, we need to ensure that all of its \textit{children} (voxels whose closest site was this one) are invalidated, adding their blocks to \textit{Indices to Clear}.


Next, we \textit{Clear Invalid}. 
The idea is to find voxels whose parents are no longer a site, and therefore need distance recomputation (see \reffig{fig:cats_clearing}).
We select a subset of the map that is within the maximum ESDF distance of the blocks in \textit{Indices to Clear}, and then check that each voxel in these blocks still has a valid parent. 
If the parent is no longer marked as a site, then the voxel distance is set to maximum, and the block index is added to \textit{Cleared Indices}.
Checking all blocks in range would seem an expensive operation but in practice, it is very efficient to check them all in parallel.


\begin{table*}[t]
\centering
\ra{1.3}
\begin{tabular}{@{}r rrrrrrr r rrrrrrr@{}}
\toprule
& \multicolumn{15}{c}{Component Runtime (ms)} \\
& \multicolumn{7}{c}{Replica~\cite{replica19arxiv}} & & \multicolumn{7}{c}{Redwood~\cite{Park2017}} \\
\cmidrule{2-8} \cmidrule{10-16}
& \multicolumn{2}{c}{Desktop} & \multicolumn{2}{c}{Laptop} & \multicolumn{2}{c}{Jetson} & & & \multicolumn{2}{c}{Desktop} & \multicolumn{2}{c}{Laptop} & \multicolumn{2}{c}{Jetson} \\
Component & nvblox & vox. & nvblox & vox. & nvblox & vox. & Speedup & & nvblox & vox. & nvblox & vox. & nvblox & vox. & Speedup \\
\midrule
ESDF & 1.9 & 163.2 & 3.6 & 291.5 & 8.4 & 231.6 & \textbf{\texttimes 63} &  & 1.5 & 29.1 & 2.6 & 46.5 & 4.2 & 38.7 & \textbf{\texttimes  16} \\
TSDF & 0.4 &  -  & 0.6 &  -  & 1.6 &  -  & \textbf{\texttimes 174} &  & 0.2 &  -  & 0.2 &  -  & 0.5 &  -  & \textbf{\texttimes 177} \\
Color & 1.7 &  -  & 2.5 &  -  & 4.2 &  -  &  -  &  & 1.1 &  -  & 1.6 &  -  & 2.4 &  -  &  -  \\
TSDF+Color & 2.1 & 86.7 & 3.2 & 106.6 & 5.8 & 226.7 & \textbf{\texttimes 38} &  & 1.3 & 38.4 & 1.8 & 33.6 & 2.9 & 76.7 & \textbf{\texttimes 25} \\
Mesh & 1.6 & 6.2 & 4.0 & 12.0 & 12.3 & 15.4 & \textbf{\texttimes 3} &  & 0.6 & 12.7 & 1.5 & 15.8 & 2.7 & 23.0 & \textbf{\texttimes 13} \\
\bottomrule
\end{tabular}
\caption{Timings for various components of \textit{nvblox} and voxblox during reconstruction of the Replica~\cite{replica19arxiv} and Redwood~\cite{Park2017} datasets at 5 cm resolution. Timings are averaged over 8 sequences for Replica, and 5 sequences for Redwood (see Sec.~\ref{sec:benchmarking} for details). \textit{Speedup} is how many times faster \textit{nvblox} is than voxblox. Some values are missing as voxblox does not separate TSDF and color integration; because of this, the TSDF speedup is assuming \textit{only} surface integration in \textit{nvblox} vs. both surface and color for voxblox (which is relevant in colorless scenarios like LiDAR integration). }
\label{tab:timings}
\end{table*}

Finally, the \textit{Lower ESDF} stage (so called since it exclusively \textit{lowers} the ESDF voxel distance) consists of two steps in a loop.
The first is \textit{Sweeping Within a Block}. This sweeps once in each axis direction in each \textit{VoxelBlock} in parallel.
This is similar to the PBA~\cite{cao2010parallel} approach, except confined to the \textit{VoxelBlock} boundaries.
\reffig{fig:cats_xplus} shows the first positive sweep within the block: each neighbor in the $X+$ direction is updated if there is a shorter distance to the site through that direction.
We then repeat the process in $X-$, $Y+$, $Y-$, $Z+$ and $Z-$. 
This is done over all affected \textit{VoxelBlocks} in a single kernel call, and at the end of the 6 sweeps, the distances within a \textit{VoxelBlock} are correct, given the current values on each \textit{VoxelBlock}'s boundaries.

We now reconcile the differences between \textit{VoxelBlocks} by \textit{Updating Neighbor Blocks} by communicating across block boundaries.
Values are propagated from the edges of one block to another if there is a shorter distance to a site through the neighboring block (see \reffig{fig:cats_border}).
%
If the last stage communicated over \textit{VoxelBlock} borders, those blocks require another sweep, to propagate the communicated distance \textit{within} the affected block.
We repeat the sweep-neighbor update loop until no more blocks can be updated.

\begin{figure}
    \centering
    \includegraphics[width=0.95\columnwidth]{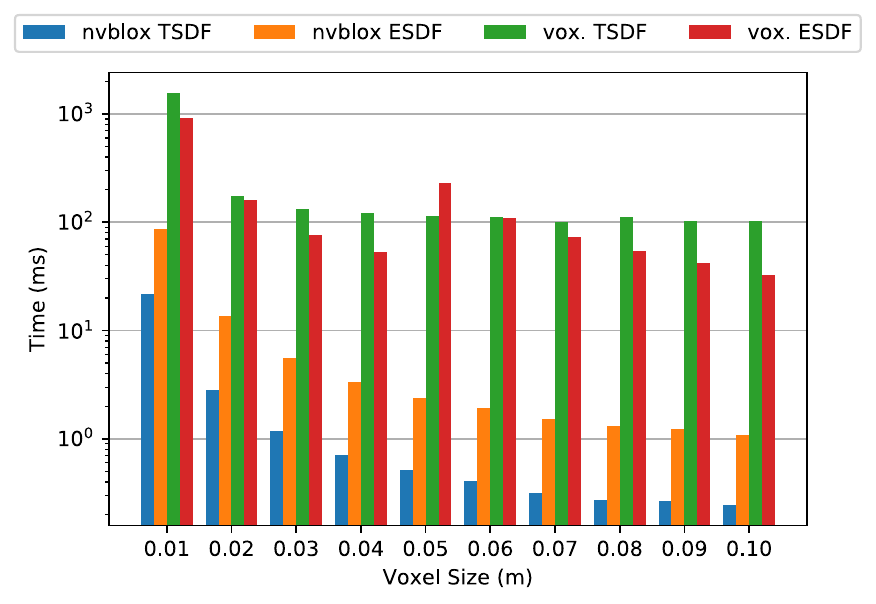}
    \caption{\ac{TSDF} and \ac{ESDF} computation times with for various voxel sizes for voxblox~\cite{oleynikova2017voxblox} and \textit{nvblox}. Timings are generated using Replica sequences~\cite{replica19arxiv} on the \textit{Laptop} compute platform.}
    \label{fig:timings_vs_resolution}
    \vspace{-5mm}
\end{figure}

\begin{figure}
    \centering
    \includegraphics[width=0.95\columnwidth]{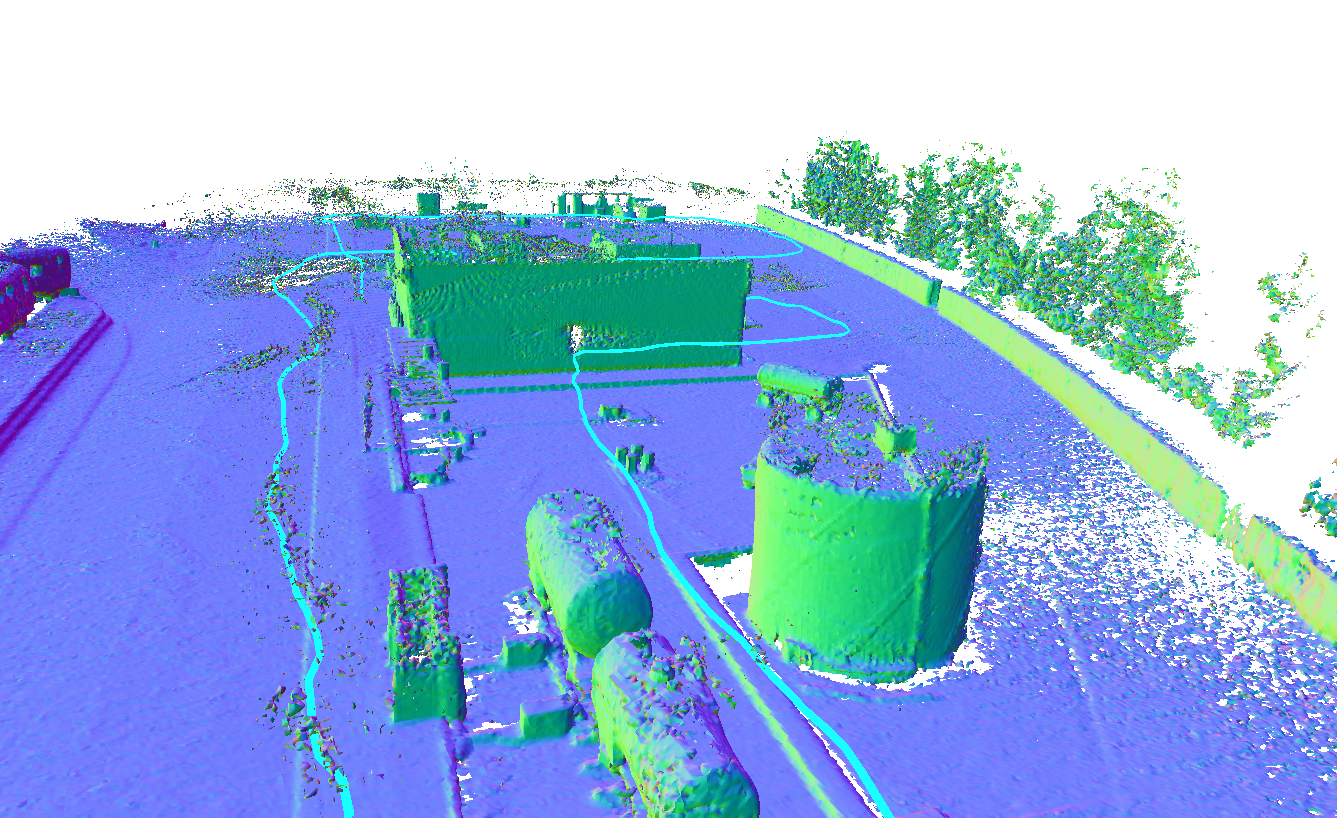}
    \caption{An example of a reconstructed mesh from a large-scale flying robot dataset. The integration time is less than 7 ms for a 64-beam LiDAR with 25 meter maximum range at 10 cm resolution on a laptop, and less than 20 milliseconds on the Jetson, allowing us to run at at least $2\times$ real time.}
    \label{fig:voxgraph_dataset}
    \vspace{-5mm}
\end{figure}

\begin{table*}[t!]
\centering
\ra{1.3}
\begin{tabular}{@{}rr rrrr r rrrr@{}}
\toprule
& & \multicolumn{3}{c}{Median ESDF Error (\textit{m})} & & \multicolumn{3}{c}{ESDF Runtime (\textit{ms}) (speedup)} \\
\cmidrule{3-5} \cmidrule{7-9}
Dataset & Sequence & nvblox & voxblox & Fiesta & & nvblox & voxblox & Fiesta \\
\midrule
Redwood~[3] & apartment & \textbf{0.04} & 0.06 & 0.05 & & 1.7 (\textbf{\texttimes 14}) & 25 (\textbf{\texttimes 1}) & 5.5 (\textbf{\texttimes 5}) \\
Redwood~[3] & bedroom & \textbf{0.02} & 0.05 & 0.03 & & 1.4 (\textbf{\texttimes 15}) & 22 (\textbf{\texttimes 1}) & 3.4 (\textbf{\texttimes 6}) \\
Redwood~[3] & boardroom & \textbf{0.06} & 0.08 & \textbf{0.06} & & 1.7 (\textbf{\texttimes 17}) & 30 (\textbf{\texttimes 1}) & 4.0 (\textbf{\texttimes 8}) \\
Redwood~[3] & lobby & 0.10 & 0.10 & \textbf{0.08} & & 2.1 (\textbf{\texttimes 16}) & 34 (\textbf{\texttimes 1}) & 5.2 (\textbf{\texttimes 7}) \\
Redwood~[3] & loft & \textbf{0.04} & 0.08 & \textbf{0.04} & & 1.8 (\textbf{\texttimes 26}) & 48 (\textbf{\texttimes 1}) & 8.4 (\textbf{\texttimes 6}) \\
Cow and lady~[4] & - & 0.09 & \textbf{0.06} & 0.07 & & 2.8 (\textbf{\texttimes 68}) & 190 (\textbf{\texttimes 1}) & 52 (\textbf{\texttimes 4}) \\
\midrule
Average & & 0.06 & 0.07 & 0.06 & & 1.9 (\textbf{\texttimes 31}) & 58 (\textbf{\texttimes 1}) & 13 (\textbf{\texttimes 4})  \\
\bottomrule
\end{tabular}
\caption{Accuracy and runtime performance of incremental ESDF generation of the proposed system compared to baselines: voxblox~\cite{oleynikova2017voxblox}, and Fiesta~\cite{han2019fiesta}. The systems are compared on the Redwood~\cite{Park2017} and Cow and Lady~\cite{oleynikova2017voxblox} datasets (see Sec.~\ref{sec:benchmarking} for details). \textit{Speedup} is the runtime performance increase over voxblox for both methods. On average, \textit{nvblox} is $7\times$ faster than Fiesta and $31\times$ faster than voxblox.}
\label{tab:esdf_timings}
\vspace{-5mm}
\end{table*}

\section{Experiments}

In this section, we aim to validate the central claim of our paper: that \textit{nvblox} improves the state-of-the-art in volumetric mapping for robot path planning in terms of run-time performance, without compromising the accuracy of the reconstructed distance field (\ac{ESDF}). We report timings on 3 different platforms: a desktop computer with an Intel i9 CPU and NVIDIA RTX3090 Ti GPU (Desktop), a laptop computer with an Intel i7 CPU and a RTX3000 Mobile GPU (Laptop), and a Jetson Xavier AGX (Jetson). 

\subsection{Whole System Timings}
\label{sec:benchmarking}

We evaluate the performance of various modules of \textit{nvblox} on the Replica Dataset~\cite{replica19arxiv}, which provides photorealistic renderings of synthetic rooms, and the Redwood dataset~\cite{Park2017} which are real scans of several environments using a consumer depth camera. For Replica, we use the sequences generated in~\cite{sucar2021imap}.

Table~\ref{tab:timings} shows timings for our system's modules; \ac{TSDF} fusion, color fusion, incremental \ac{ESDF} update, and incremental meshing. We perform \ac{ESDF} generation and meshing every 4 frames. Timings are averaged across 8 Replica Dataset sequences and 5 Redwood Dataset sequences. 

When compared to our previous work, voxblox~\cite{oleynikova2017voxblox}, which runs on the CPU, we see significant speed-ups in all modules of the system. 
In the case of \ac{TSDF} and color this is a well-known result, as the GPU has been used to accelerate \ac{TSDF} mapping since KinectFusion~\cite{izadi2011kinectfusion}.
We show two additional findings. These speed-ups are also achievable on an embedded GPU. Furthermore, similar speed-ups are available for incremental ESDF calculation, which we describe below.

\subsection{ESDF Timings}

We aim to validate our claim of improving the state-of-the-art in incremental \ac{ESDF} calculation. 
We compare \textit{nvblox} against voxblox~\cite{oleynikova2017voxblox}, and Fiesta~\cite{han2019fiesta}, a recent and more performant algorithm. 
Table~\ref{tab:esdf_timings} shows timings and \ac{ESDF} accuracy for our \textit{Desktop} system. 
The \ac{ESDF} error is calculated as the median absolute voxel-wise error between the reconstructed \ac{ESDF} and a voxelized \ac{ESDF} ground-truth. The ground-truth is generated by computing the distance between the reconstructed voxel centers, and the dataset-supplied ground-truth surface. 
Table~\ref{tab:esdf_timings} shows a significant speedup of $31\times$ with respect to voxblox and $7\times$ with respect to Fiesta. 
Furthermore, the experiments show that this speed-up does not come at the cost of reduced accuracy.

\subsection{Resolution Scaling}

The relationship between map resolution and \ac{TSDF} and \ac{ESDF} computation time is critical because map resolution is often adjusted to meet performance limitations. 
We run \textit{nvblox} and voxblox~\cite{oleynikova2017voxblox} on the \textit{office0} Replica sequence for various resolutions. Fig.~\ref{fig:timings_vs_resolution} shows the results of this experiment. Even at high resolution ($1\,\text{cm}$ for \ac{TSDF} and $2\,\text{cm}$ for \ac{ESDF}) \textit{nvblox} performs computations faster than voxblox running at the lowest tested resolution $10\,\text{cm}$. This speed-up is likely to enable robotic applications requiring higher precision 3D perception.

\subsection{Query Timings}

The primary purpose of mapping in a typical robotic system is to provide collision information to path-planning modules.
For many optimization or sampling-based planners, querying for collisions can constitute a significant portion of the total computational cost.
Because these queries are often required in batch, performing these queries on the GPU allows us to take advantage of parallelization, and enables GPU-based path planners, like in \cite{sundaralingam2023curobo} and \cite{pantic2023obstacle}. 
A query in \textit{nvblox} takes a collection of 3D points and returns their distances to the closest surface and optionally the distance field gradient, by performing an \ac{ESDF} lookup on the GPU. 
Table~\ref{tab:queries} shows query rates in giga-queries-per-second for an NVIDIA GeForce 3090 Ti as well as a Jetson AGX. 
The table shows the results for spatially correlated (cor.) and uncorrelated (uncor.) sampling. 
In correlated sampling, adjacent queries are more likely to fall in the same \textit{VoxelBlock}, leading to coalesced memory access on the GPU and higher query rates. 
This is typically the case for robotic use cases where the queries are spatially correlated, for example, clustered around the robot's current location.

\begin{table}[t]
\centering
\begin{tabular}{lrr r rr r rr}
\toprule
{} & \multicolumn{8}{c}{$10^9$ Queries per Second}  \\
\cmidrule{2-9}
{} & \multicolumn{2}{c}{Desktop} & & \multicolumn{2}{c}{Laptop} & & \multicolumn{2}{c}{Jetson} \\
\cmidrule{2-3} \cmidrule{5-6} \cmidrule{8-9} 
Dataset & cor. & uncor. & & cor. & uncor. & & cor. & uncor. \\
\midrule
Redwood & 6.2 & 3.3 & & 1.7 & 1.3 & & 0.8 & 0.5\\
Sun3D & 7.3 & 3.3 & & 1.8 & 1.1 & & 0.7 & 0.3\\
\bottomrule
\end{tabular}
\caption{The number of distance giga-queries per second delivered by \textit{nvblox}, averaged over several sequences of the Redwood~\cite{Park2017} and Sun3D datasets~\cite{xiao2013sun3d}. Even on the Jetson, uncorrelated queries take only 3 \textit{nanoseconds} per point.\vspace{-0em}}
\label{tab:queries}
\end{table}

\begin{figure}
    \centering
    \includegraphics[width=0.95\columnwidth]{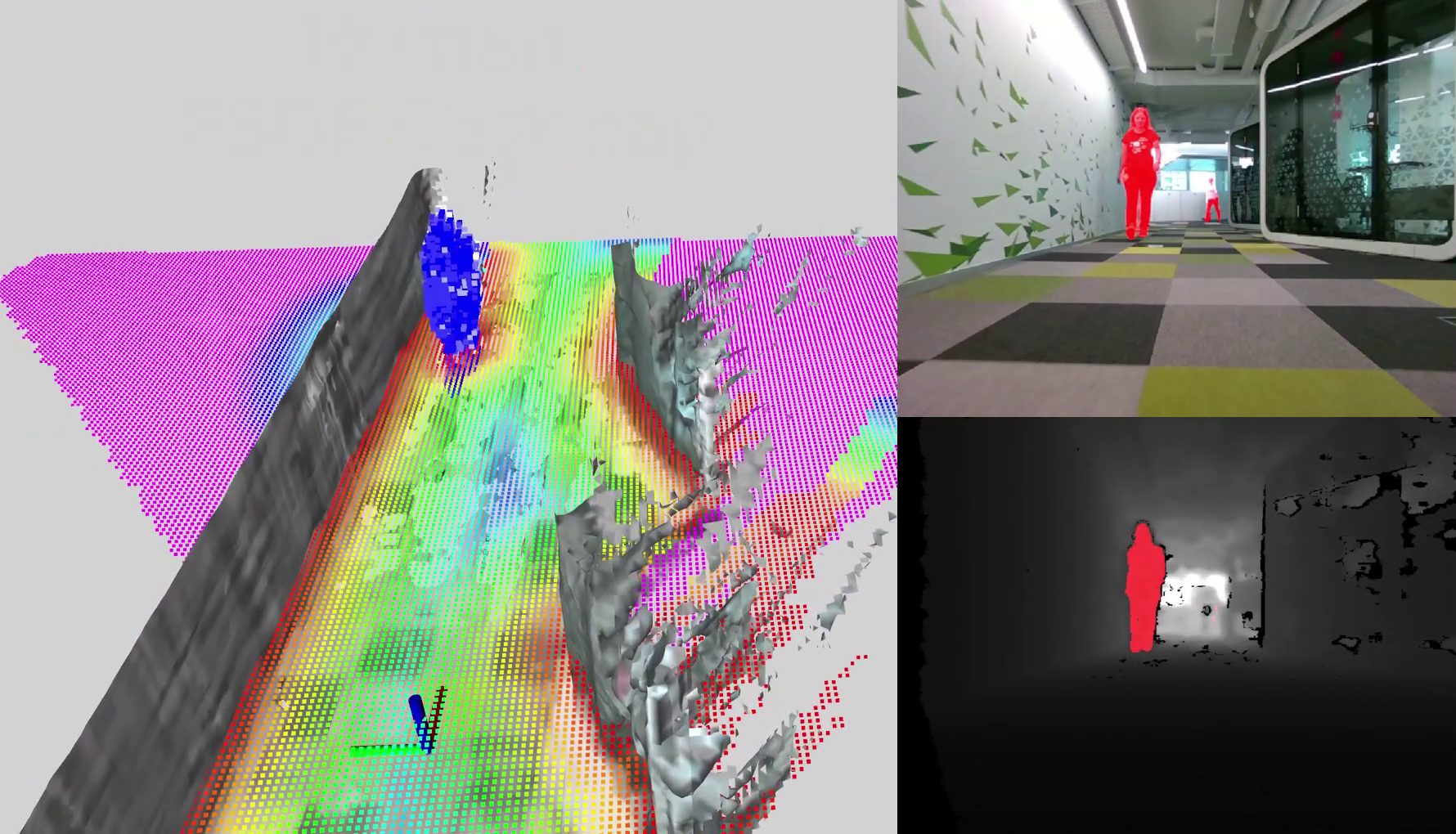}
    \caption{An example of \textit{nvblox} on a ground robot in an office building. The input color and depth images, as well as the semantic segmentation are shown on the right. On the left is the resulting reconstruction showing the reconstructed mesh, a slice through the distance field, as well as occupancy-probability voxels representing the reconstructed human (in blue).
    }
    \label{fig:human_segmentation}
    \vspace{-5mm}
\end{figure}

\subsection{Application examples}
To demonstrate the wide utility of \textit{nvblox} at various problem scales we show examples of its use on flying robots, robot arms, and mobile ground robots.
\reffig{fig:voxgraph_dataset} shows the results from a dataset collected with a drone~\cite{reijgwart2019voxgraph} equipped with a 64-beam Ouster OS1 LiDAR.
We use Fast-LIO~\cite{xu2021fast} as a pose estimator and integrate the LiDAR up to a range of 25 meters at a resolution of 5 cm, which requires less than 7 ms per LiDAR scan on a Laptop.
This is an example of \textit{nvblox}'s suitability for large-scale outdoor scenarios, where the resulting 3D ESDF can be used for both global~\cite{oleynikova2018sparse} and local planning~\cite{oleynikova2020open}.
The \textit{nvblox} library has also been used on-board a different flying robot to enable Riemannian Motion Policies which allow reactive navigation at kHz rates~\cite{pantic2023obstacle}.

\textit{Nvblox} is suitable for small-scale problems as well, as shown for high-rate adaptive planning for robot arms in CuRobo~\cite{sundaralingam2023curobo}, where the authors take advantage of \textit{nvblox}'s fast query speeds directly on GPU to sample more trajectory candidates than previously possible. In general, \textit{nvblox} is useful not only because it is faster than existing methods, but also because all data is already stored on the GPU. This enables integration with other GPU-accelerated methods, without requiring CPU-GPU memory transfers.

Lastly, we show an image from a robot in an office environment~\reffig{fig:human_segmentation}. 
In this example, we use PeopleSemSegnet\footnote[5]{\rurl{catalog.ngc.nvidia.com/orgs/nvidia/teams/tao/models/peoplesemsegnet}} to segment the reconstruction into the static environment and dynamic elements (e.g. humans).
Depth image segments belonging to the human class are fed into a 3D occupancy grid using a separate \textit{OccupancyLayer} in \textit{nvblox}.
The result is a two-part reconstruction, where dynamic parts of the scene decay over time, but the static parts of the scene are accurately reconstructed using \ac{TSDF} fusion.

\section{Conclusion} 
\label{sec:conclusion}

In conclusion, we introduce \textit{nvblox}, a library for volumetric mapping on the GPU. 
The library fills a gap between CPU-based volumetric mapping systems for robots, which are computationally limited, and GPU-based systems that typically omit features that are important for robotics use-cases. 
As part of the toolbox we include a novel incremental, GPU-accelerated method for computing the \ac{ESDF}. 
The system is optimized for operation on both discrete and embedded GPUs. 
We provide experiments demonstrating that \textit{nvblox} is significantly faster both in mapping and distance field computation, as well as at query time, than other state-of-the-art approaches.


\printbibliography


\end{document}